\documentclass[supp]{tlp}

\usepackage{supp}

\usepackage[utf8]{inputenc}
\usepackage{hyperref}
\usepackage{amssymb}
\usepackage{amsmath}
\usepackage{url}
\usepackage{graphicx}
\usepackage{algorithm2e}
\usepackage{mdwlist}

\usepackage{times} 

\newcommand{\tr}{$\mathcal{TR}$}

\newcommand{\name}{$\mathcal{TR}^{ev}$}
\newcommand{\trev}{$\mathcal{TR}^{ev}$}

\newcommand{\pa}{\mathtt{path}}
\newcommand{\resp}[1]{\mathbf{r}(#1)}
\newcommand{\occ}[1]{\mathbf{o}(#1)}

\newcommand{\ex}[1]{\mathtt{exp}_{M}(#1)}

\newcommand{\fu}{\mathit{choice}}
\newcommand{\paempty}{\epsilon}

\newcommand{\trans}[1]{$ $\!^{#1}\!\!\!\to\!}
\newcommand{\hyp}{\lozenge}

\newcommand{\extended}[1]{}

\newcommand{\seqDash}{\textendash}

\newcommand{\comment}[1]{}

\newtheorem{definition}{Definition}
\newtheorem{lemma}{Lemma}

\newtheorem{theorem}{Theorem}
\newtheorem{example}{Example}

\begin{document}

\title{Transaction Logic with (Complex) Events}

\author[Ana Sofia Gomes and Jos\'e J\'ulio Alferes]{Ana Sofia Gomes and Jos\'e J\'ulio Alferes\thanks{The authors thank Michael Kifer for the valuable discussions in a preliminary version of this work. The first author was supported by the grant SFRH/BD/64038/2009 and by project ERRO (PTDC/EIA-CCO/121823/2010). The second author was supported by project ASPEN PTDC/EIA-CCO/110921/2009} \\
	CENTRIA - Dep. de Inform\'atica, Faculdade Ci\^{e}ncias e Tecnologias \\
       Universidade Nova de Lisboa}

\pubauthor{A. S. Gomes and J. J. Alferes}
\jurl{xxxxxx}
\pubdate{xxx}

\maketitle

\begin{abstract}
This work deals with the problem of combining reactive features, such as the ability to respond to events and define complex events, with the execution of transactions over general Knowledge Bases (KBs).

With this as goal, we build on Transaction Logic (\tr{}), a logic precisely designed to model and execute transactions in KBs defined by arbitrary logic theories. In it, transactions are written in a logic-programming style, by  combining primitive update operations over a general KB, with the usual logic programming connectives and some additional connectives e.g. to express sequence of actions. 
While \tr{} is a natural choice to deal with transactions, it remains the question whether  \tr{} can be used to express complex events, but also to deal simultaneously with the detection of complex events and the execution of transactions.  
In this paper we show that the former is possible while the latter is not. For that, we start by illustrating how \tr{} can express complex events, and in particular, how SNOOP event expressions can be translated in the logic.
Afterwards, we show why \tr{} fails to deal with the two issues together, and to solve the intended problem propose Transaction Logic with Events, its syntax, model theory and executional  semantics.
The achieved solution is a non-monotonic  extension of \tr{}, which guarantees that every complex event detected in a transaction is necessarily responded. 

To appear in Theory and Practice of Logic Programming (TPLP)
\end{abstract}

 \vspace{-.3cm}
\section{Introduction}
Reactivity stands for the ability to detect complex changes (also denoted as events) in the environment and react automatically to them according to some pre-defined rules. This is a pre-requisite of many real-world applications, such as web-services providing different services depending on external information, multi-agent systems adapting their knowledge and actions according to the  changes in the environment, or monitoring systems reacting to information detected by their sensors and issuing actions automatically in response to it. In reactive systems, e.g. in those based on Event-Condition-Action (ECA) languages \cite{AlferesBB11,BryEP06,ChomickiLN03}, the reaction triggered by the detection of a complex event may itself be a complex action, formed e.g. by the sequencial execution of several basic actions. Moreover, we sustain that sometimes reactive systems are also required to execute \emph{transactions} in response to events.  For example, consider an airline web-service scenario where an external event arrives stating that a partner airline is on strike for a given time period. Then, the airline must address this event by e.g. rescheduling flights with alternative partners or refund tickets for passengers who do not accept the changes. Clearly, some transactional properties regarding these actions must be ensured: viz. it can never be the case that a passenger is simultaneously not refunded nor have an alternative flight; or that she is completely refunded and has a rescheduled flight.

Although the possibility of executing transactions is of crucial importance in many of today's systems, and a must e.g. in  database systems, most reactive languages do not deal with it.
Some exceptions exist, but are either completely procedural and thus lack from a clear declarative semantics (as e.g. in \cite{PapamarkosPW06}), or have a strong limitation on the expressivity of either the actions or events (as e.g. in \cite{Zaniolo95,LausenLM98a}). 

In this paper we propose Transaction Logic with Events, \trev{}, an extension of \tr{}~\cite{BonnerK93} integrating the ability to reason and execute transactions over very general forms of KBs, with the ability to detect complex events. For this, after a brief overview of \tr{}, we show how it can be used to express and reason about complex events, and in particular, how it can express most SNOOP event operators~\cite{snoopIB} (Section~\ref{sec:events-tr}). We proceed by showing why \tr{} alone is not able to deal with both the detection of complex events and the execution of transactions, and, in particular, why it does not guarantee that all complex events detected during the execution of a transaction are responded within that execution. For solving this problem, we define \trev{}, its language and model theory (Section~\ref{sec:trev:semantics}), as well as  its executional semantics (Section~\ref{sec:entailment}). 

\vspace{-.3cm}
\section{Using \tr{} to express complex events}
\label{sec:events-tr}

In this section we briefly recall \tr{}'s syntax and semantics with minor syntactic changes from the original, to help distinguish between actions and event occurrences, something that is useful ahead in the paper when extending \tr{} to deal with reactive features and complex events. 

Atoms in \tr{} have the form $p(t_1,\ldots,t_n)$ where $p$ is a predicate symbol and $t_i$'s are terms (variables, constants, function terms). 
For simplicity, and without loss of generality \cite{BonnerK98b}, we consider Herbrand instantiations, as usual. 
To build complex formulas, \tr{} uses the classical connectives $\land, \lor, \neg, \leftarrow$ and the connectives $\otimes, \hyp$ denoting serial conjunction and hypothetical execution.
Informally,  $\phi \otimes \psi$ is an action composed of an execution of $\phi$ followed by an execution of $\psi$; and $\hyp \phi$ tests if $\phi$ can be executed without materializing the changes. In general, formulas are viewed as (the execution of) transactions, where, $\phi \land \psi$ is the simultaneous execution of $\phi$ and $\psi$;  $\phi \lor \psi$  the non-deterministic choice of executing $\phi$ or $\psi$.
$\phi \leftarrow \psi$ is a \emph{rule} saying that one way to execute of $\phi$ is by executing $\psi$.
As in classical logic, $\land$ and $\leftarrow$ can be written using $\lor$ and $\neg$ (e.g.  $\phi \land \psi \equiv \neg (\neg \phi \lor \neg \psi)$). 
Finally, we also use the connective $;$ as it is useful to express common complex events. $\phi;\psi$ says that $\psi$ is true after $\phi$ but possibly interleaved with other occurrences, and it can be written in \tr{} syntax as: $\phi \otimes \pa \otimes \psi$ where $\pa \equiv (\varphi \lor \neg \varphi)$ is a tautology that holds in paths of arbitrary size \cite{BonnerK98b}.


For making possible the separation between the theory of states and updates, from the logic that combines them in transactions, \tr{} considers a pair of oracles -- $\mathcal{O}^d$ (data oracle) and $\mathcal{O}^t$ (transition oracle) -- as a parameter of the theory. These oracles are mappings that assume a set of \emph{state identifiers}. 
$\mathcal{O}^{d}$ is a mapping from state identifiers to a set of formulas that hold in that state, and $\mathcal{O}^{t}$ is a mapping from pairs of state identifiers to sets of formulas that hold in the transition of those states.
These oracles can be instantiated with a wide variety of semantics, as e.g. relational databases, well-founded semantics, action languages, etc. \cite{BonnerK93}.  For example, a relational database can be modeled by having states represented as sets of ground atomic formulas. Then, the data oracle simply returns all these formulas, i.e., $\mathcal{O}^d(D) = D$, and for each predicate $p$ in the KB, the transition oracle defines $p.ins$ and $p.del$, representing the insertion and deletion of $p$, respectively. Formally, $p.ins \in \mathcal{O}^{t}(D_1,D_2)$ iff $D_2 = D_1 \cup \{p\}$ and, $p.del \in \mathcal{O}^t(D_1,D_2)$ iff $D_2 = D_1 \backslash \{p\}$. SQL-style bulk updates can also be defined by $\mathcal{O}^{t}$.
%

%

\vspace{-.1cm}
\begin{example}[Moving objects - \tr{}]
\label{ex:intro1}
As a \tr{}'s illustration, assume the prior relational database oracles and the action $move(O,X,Y)$ defining the relocation of object $O$ from position $X$ into position $Y$.
In such a KB, states are defined using the predicates $location(O,P)$ saying that object $O$ is in position $P$, and $clear(X)$ stating that $X$ is clear to receive an object.
In \tr{}, the move (trans)action can be expressed by:
$$
\small
\begin{array}{l}
move(O,X,Y) \leftarrow location(O,X) \otimes clear(Y) \otimes localUpdt(O,X,Y) \\
localUpdt(O,X,Y) \leftarrow location(O,X).del \otimes location(O,Y).ins \otimes clear(Y).del \otimes clear(X).ins
\end{array}
$$
\end{example}

\tr{}'s theory is built upon the notion of sequences of states denoted as \emph{paths}. 
Formulas are evaluated over paths, and truth in \tr{} means \emph{execution}: a formula is said to succeed over a path, if that path represents a valid execution for that formula. 
Although not part of the original \tr{}, here paths' state transitions are labeled with information about what (atomic occurrences) happen in the transition of states.
Precisely, paths have the form $\langle D_0\trans{O_{1}}D_{2}\trans{O_{2}} \ldots\trans{O_{k}} D_k \rangle$, where $D_{i}$'s are states and $O_{i}$'s are labels (used later to annotate atomic event occurrences).
%
%

As usual, satisfaction of complex formulas is based on interpretations. These define what atoms are true in what paths, by mapping every path to a set of atoms. However, only the mappings compliant with the specified oracles are interpretations:

\begin{definition}[Interpretation]
\label{def:interpretations}
An interpretation is a mapping $M$ assigning a set of atoms (or $\top$\footnote{For not having to consider partial mappings, besides formulas, interpretations can also return the special symbol $\top$. The interested reader is referred to \cite{BonnerK93} for details.}) to every path, with the following restrictions (where $D_i$s are states, and $\varphi$ a formula):
\vspace{-.15cm}
\begin{enumerate}
\item $\varphi \in M(\langle D \rangle)$ \qquad \qquad \qquad  \qquad \qquad  \ \!\! \text{    if } $\varphi \in \mathcal{O}^{d}(D) $
\label{point-query}
\item $\{\varphi, \occ{\varphi}\} \subseteq M(\langle D_{1}\trans{\occ{\varphi}}D_{2} \rangle)$ \qquad \ \text{   if } $\varphi\in\mathcal{O}^{t}(D_{1},D_{2})$ 
\label{point-elementary}
\label{point-tech}
\end{enumerate}
\end{definition}
%
In point~\ref{point-elementary} we additionally (i.e., when compared to the original definition) force $\occ{\varphi}$ to belong to the same path where the primitive action $\varphi$ is made true by the oracle, something that later (in Section~\ref{sec:trev}) will help detect events associated with primitive actions, like ``on insert/delete''.

Next, we define operations on paths, and satisfaction of complex formulas over general paths. 



\begin{definition}[Path Splits, Subpaths and Prefixes]
\label{def:split}
Let $\pi$ be a $k$-path, i.e. a path of length $k$ of the form $\langle D_{1}\trans{O_{1}} \ldots \trans{O_{k-1}} D_{k} \rangle$.
A \emph{split} of $\pi$ is any pair of subpaths, $\pi_1$ and $\pi_2$, such that $\pi_1 = \langle D_1\trans{O_{1}} \ldots \trans{O_{i-1}} D_i\rangle$ and $\pi_2 = \langle D_i\trans{O_{i}}\ldots \trans{O_{k-1}} D_k \rangle$ for some $i$ $(1 \leq i \leq k)$. In this case, we write $\pi = \pi_1 \circ \pi_2$.

\noindent A subpath $\pi'$ of $\pi$ is any subset of states and annotations of $\pi$ where both the order of the states and their annotations is preserved.
A prefix $\pi_{1}$ of $\pi$ is any subpath of $\pi$ sharing the initial state. 
\end{definition}


\begin{definition}[\tr{} Satisfaction of Complex Formulas] 
\label{def:satev}
Let $M$ be an interpretation, $\pi$ a path and $\phi$ a formula. If $M(\pi) = \top$ then $M,\pi \models_{\mathcal{TR}} \phi$; else:
\vspace{-.2cm}
\begin{enumerate}[(iii)]
\item \textbf{Base Case:} $M,\pi \models_{\mathcal{TR}} \phi$ iff $\phi \in M(\pi)$ for every event occurrence $\phi$
\item \textbf{Negation:} $M,\pi\models_{\mathcal{TR}} \neg \phi$ iff it is not the case that $M, \pi \models_{\mathcal{TR}} \phi$
\item \textbf{Disjunction:} $M,\pi \models_{\mathcal{TR}} \phi \lor \psi$ iff $M,\pi \models_{\mathcal{TR}}\phi$ or $M,\pi\models_{\mathcal{TR}} \psi$.
\item \textbf{Serial Conjunction:} $M,\pi \models_{\mathcal{TR}} \phi \otimes \psi$ iff there exists a split $\pi_{1} \circ \pi_{2}$ of $\pi$ s.t.
$M,\pi_{1}\models_{\mathcal{TR}} \phi$ and $M,\pi_{2}\models_{\mathcal{TR}}\psi$
\item \textbf{Executional Possibility:} $M,\pi \models_{\mathcal{TR}} \hyp \phi$ iff $\pi$ is a 1-path of the form $\langle D \rangle$ for some state $D$ and $M,\pi' \models_{\mathcal{TR}} \phi$ for some path $\pi'$ that begins at $D$.
\end{enumerate}
\end{definition}

Models and logical entailment are defined as usual. An interpretation models/satisfies a set of rules if each rule is satisfied in every possible path, and an interpretation models a rule in a path, if whenever it satisfies the antecedent, it also satisfies the consequent. 

\vspace{-.1cm}
\begin{definition}[Models, and Logical Entailment]
\label{model-tr}
An interpretation $M$ is a \emph{model} of a formula $\phi$ iff for every path $\pi$, $M,\pi \models_{\mathcal{TR}} \phi$.
\noindent $M$ is a model of a set of rules $P$ (denoted $M \models_{\mathcal{TR}} P$) iff it is a model of every rule in $P$.

\noindent $\phi$ is said to logically entail another formula $\psi$ iff every model of $\phi$ is also a model of $\psi$.
\end{definition}

Logical entailment is useful to define general equivalence and implication of formulas that express properties like ``transaction $\phi$ is equivalent to transaction $\psi$'' or ``whenever transaction $\psi$ is executed, $\psi'$ is also executed''.
Moreover, if instead of transactions, we view the propositions as representing event occurrences, this entailment can be used to express complex events. 
For instance, imagine  we want to state a complex event $alarm$, e.g. triggered whenever event $ev_1$ occurs after both $ev_2$ and $ev_3$ occur simultaneously.
This can be expressed in \tr{} as: 
\begin{equation}
\small
\label{ex:alarm}
\occ{alarm} \leftarrow (\occ{e_2} \land \occ{e_3}) ; \occ{e_1}
\end{equation}
In every model of this formula, whenever there is a (sub)path where both $\occ{e_2}$ and $\occ{e_3}$ are true, followed by a (sub)path where $ \occ{e_1}$ holds, then $\occ{alarm}$ is true in the \emph{whole} path.


Other complex event definitions are possible, and in fact we can encode most of SNOOP \cite{snoopIB} operators in \tr{}.
This is shown in Theorem~\ref{theorem:snoop} where, for a given history of past event occurrences, we prove that if an event expression is true in SNOOP, then there is a translation into a \tr{} formula which is also true in that history.
Since a SNOOP history is a set of atomic events associated with discrete points in time, the first step is to build a \tr{} path expressing such history. We construct it as a sequence of state identifiers labeled with time, where time point $i$ takes place in the transition of states $\langle s_i,s_{i+1} \rangle$, and only consider interpretations $M$ over such a path that are \emph{compatible} with SNOOP's history, i.e. such that, for every atomic event that is true in a time $i$, $M$ makes the same event true in the path $\langle s_i,s_{i+1}\rangle$.

\begin{theorem}[SNOOP Algebra and \tr{}]
\label{theorem:snoop}
Let $E$ be a \emph{SNOOP} algebra expression without periodic and aperiodic operators, $H$ be a history containing the set of all SNOOP primitive events $e^{i}_{j}[t_1]$ that have occurred over the time interval $t_1,t_{max}$, and $\langle s_1, \ldots s_{max+1} \rangle$ be a path with size $t_{max}-t_1+1$.
Let $\tau$ be the following function: 
\vspace{-.2cm}
\[\small
\begin{array}{l@{\ }l}
$\textbf{Primitive:}$ & \tau(E) = \occ{E}$  where $E$ is a primitive event$\\
$\textbf{Sequence:}$ &\tau(E_1;E_2) = \tau(E_1)\otimes\pa\otimes\tau(E_2)\\
$\textbf{Or:}$ & \tau(E_1 \triangledown E_2) = \tau(E_1) \lor \tau(E_2)\\
$\textbf{AND:}$ & \tau(E_1 \triangle E_2) =  [(\tau(E_1)\otimes \pa) \land (\pa \otimes \tau(E_2))] \lor
[(\tau(E_2)\otimes \pa) \land (\pa \otimes \tau(E_1))]\\
$\textbf{NOT:}$ & \tau(\neg (E_3)[E_1,E_2]) =  \tau(E_1)\otimes \neg \tau(E_3) \otimes\tau(E_2)
\end{array}\]
Then,  $[t_i,t_f] \in E[H] \Rightarrow \forall$\emph{M compatible with $H$},  $M, \langle s_{t_i},\ldots,s_{t_{f+1}} \rangle \models_{\mathcal{TR}} \tau(E)$,
where, cf. \cite{snoopIB}, $E[H]$ is the set of time intervals $(t_i,t_f)$ where $E$ occurs over $H$ in an unrestricted context, and where $M$ is compatible with $H$ if, for each $e^{i}_{j}[t_i] \in H$: $M, \langle s_{t_i} , s_{t_{i+1}} \rangle \models_{\mathcal{TR}} \occ{e_j}$.
\end{theorem}

Besides the  logical entailment, \tr{} also provides the notion of executional entailment for reasoning about properties of a \emph{specific} execution path.
\begin{definition}[Executional Entailment]
\label{def:entailment}
Let $P$ be a set of rules, $\phi$ a formula, and $D_0\trans{O_{1}}\ldots\trans{O_{n}} D_n$ a path.\\
$P, (D_0\trans{O_{1}}\ldots\trans{O_{n}} D_n) \models \phi$ ($\star$)
iff for every model $M$ of $P$, $M,\langle D_0\trans{O_{1}}\ldots\trans{O_{n}} D_n \rangle \models \phi$. 

\noindent Additionally, $P,D_0\seqDash \models \phi$
holds, if there is a path $D_0\trans{O_{1}}\ldots\trans{O_{n}} D_n$ that makes ($\star$) true.
\end{definition}

$P, (D_0\trans{O_{1}}\ldots\trans{O_{n}} D_n) \models \phi$ says that a successful execution of transaction $\phi$ respecting the rules in $P$, can change the KB from state $D_0$ into $D_n$ with a sequence of occurrences $O_{1},\ldots,O_{n}$. 
E.g., in the Example~\ref{ex:intro1} (with obvious abbreviations),  the statement $P,(\{ cl(t), l(c,o) \}\trans{\occ{l(c,o).del}} \{ cl(t) \}\trans{\occ{l(c,t).ins}}\{  cl(t), l(c,t)\} \trans{\occ{cl(t).del}} \{ l(c,t) \} \trans{\occ{cl(o).ins}}\{ l(c,t), cl(o) \}) \models  move(c,o,t)$ means that a possible result of executing the transaction $move(c,oven,table)$ starting in the state $\{ clear(table), loc(c,oven) \}$ is the path with those 5 states, ending in $\{ loc(c,table), clear(oven)\}$.

This entailment has a corresponding proof theory \cite{BonnerK93} which, for a subset of \tr{}, is capable of \emph{constructing} such a path given a program, a \tr{} formula, and an initial state. I.e. a path where the formula can be executed. If no such path exists, then the transaction fails, and nothing is built after the initial state.

\vspace{-.3cm}
\section{\trev{}: combining the execution of transactions with complex event detection}
\label{sec:trev}

Reactive languages need to express behaviors like:
``on $alarm$ do action $a_1$ followed by action $a_2$", where the actions $a_1 \otimes a_2$ may define a transaction, and $alarm$ is e.g. the complex event in (\ref{ex:alarm}).
Clearly, \tr{} can individually express and reason about transaction $a_1 \otimes a_2$, and its complex event. So, the question is whether it can deal with both simultaneously. For that, two important issues must be tackled: 1) how to model the triggering behavior of reactive systems, where the occurrence of an event drives the execution of a transaction in its response; 2) how to model the transaction behavior that prevents transactions to commit until all occurring events are responded. 

Regarding 1), \cite{BonnerKC93} shows that simple events can be triggered in \tr{} as: 
\vspace{-.1cm}
\begin{equation}
\small
\label{eq:example-tr}
\begin{array}{l}
p \leftarrow body \otimes ev \\
ev \leftarrow  \resp{ev}
\end{array}
\vspace{-.1cm}
\end{equation}
With such rules, in all paths that make $p$ true (i.e., in all executions of transaction $p$) the event $ev$ is triggered/fired (after the execution of some arbitrary $body$), and $ev$'s response, $\resp{ev}$, is executed. Note that, both $\resp{ev}$ and $body$ can be defined as arbitrary formulas.

But, this is just a very simple and specific type of event: atomic events that are explicitly triggered by a transaction defined in the program.
In general, atomic events can also arrive as external events, or because some primitive action is executed in a path (e.g. as the database triggers - ``on insert/on delete"). Triggering external events in \tr{}  can be done by considering the paths that make the external event true. E.g., if one wants to respond to an external event $ev$ from an initial state, all we need to do is find the paths $\pi$ starting in that state, s.t. $P, \pi \models ev$, where $P$ includes the last rule from (\ref{eq:example-tr}) plus the rules defining $ev$'s response.

The occurrences of primitive actions can be tackled by  Point~\ref{point-elementary} of Def.~\ref{def:interpretations}, and the occurrence of complex events can be defined as prescribed in Section~\ref{sec:events-tr}. However, the above approach of \cite{BonnerKC93} does not  help for driving the execution of an event response when such occurrences become true.
%
%
%
%
For instance, the ECA-rule before could be stated as:   
$$
\label{ex:eca}
\vspace{-.1cm}
\small
\begin{array}{l}
\occ{alarm} \leftarrow (\occ{e_2} \land \occ{e_3}) ; \occ{e_1}  \\
\resp{alarm} \leftarrow a_1 \otimes a_2
\end{array}
$$ 
But this does not drive the execution of $\resp{alarm}$ when $\occ{alarm}$ holds; one has further to force that whenever $\occ{alarm}$ holds, $\resp{alarm}$ must be made true subsequently. Of course, adding a rule $\resp{alarm} \leftarrow \occ{alarm}$ would not work: such rule would only state that, one alternative way to satisfy the response of alarm is to make its occurrence true. And for that, it would be enough to satisfy $\occ{alarm}$ to make $\resp{alarm}$ true, which is not what is intended.


Clearly, this combination implies two different types of formulas with two very different behaviors: the \emph{detection} of events which are tested for occurrence w.r.t. a past history; and the \emph{execution} of transactions as a response to them, which intends to construct paths where formulas can succeed respecting transactional properties.
This has to be reflected in the semantics and these formulas should be evaluated differently accordingly to their nature. 

Regarding 2), as in database triggers, transaction's execution must depend on the events triggered. Viz., an event occurring during a transaction execution can delay that transaction to commit/succeed until the event response is successfully executed, and the failure of such response should imply the failure of the whole transaction. 
%
Encoding this behavior requires that, if an event occurs during a transaction, then its execution needs to be \emph{expanded} with the event response.
Additionally, this also precludes transactions to succeed in paths where an event occurs and is not responded (even if the transaction would succeed in that path if the event did not existed).

For addressing these issues, below we define \trev{}. This extension of \tr{} evaluates event formulas and transaction formulas differently, using two distinct relations (respectively $\models_{\mathcal{TR}}$ and $\models$), and occurrences and responses are syntactic represented w.r.t. a given event name $e$, as $\occ{e}$ and $\resp{e}$, respectively.
In this context, $\models$ requires transactions to be satisfied in expanded paths, where every occurring event (made true by $\models_{\mathcal{TR}}$) is properly responded.

\vspace{-.2cm}
\subsection{\trev{} Syntax and Model Theory}
\label{sec:trev:semantics}
To make possible a different evaluation of events and transactions, predicates in \trev{} are  partitioned into transaction names ($\mathcal{P}_{t}$), event names ($\mathcal{P}_{e}$), and oracle primitives ($\mathcal{P}_{\mathcal{O}}$) and, as with \tr{}, we work with the Herbrand instantiation of the language.

Formulas in \trev{} are partitioned into transaction formulas and event formulas.
\emph{Event formulas} denote formulas meant to be \emph{detected} and are either an event occurrence, or an expression defined inductively as $\neg \phi$, $\phi \land \psi$, $\phi \lor \psi$, $\phi \otimes \psi$, or $\phi ; \psi$ where $\phi$ and $\psi$ are event formulas. An \emph{event occurrence} is of the form $\occ{\varphi}$ s.t. $\varphi \in \mathcal{P}_e$ or $\varphi \in \mathcal{P}_\mathcal{O}$. 
Note that, we preclude the usage of  $\hyp$ in event formulas, as it would make little sense to detect occurrences based on what could possibly be executed.

\emph{Transaction formulas} are formulas that can be \emph{executed}, and are either a transaction atom, or an expression defined inductively as $\neg \phi$, $\hyp \phi$, $\phi \land \psi$, $\phi \lor \psi$, or $\phi \otimes \psi$.  A  \emph{transaction atom} is either a transaction name (in $\mathcal{P}_t$), an oracle defined primitive (in $\mathcal{P}_\mathcal{O}$), the response to an event ($\resp{\varphi}$ where $\varphi \in \mathcal{P}_{\mathcal{O}} \cup \mathcal{P}_e$), or an event name (in $\mathcal{P}_e$) The latter corresponds to the (trans)action of \emph{explicitly} triggering an event directly in a transaction as in (\ref{eq:example-tr}) or as an external event. As we shall see (Def.~\ref{def:sat}) explicitly triggering an event changes the path of execution (by asserting the information that the event has happened in the current state) and, as such, is different from simply inferring (or detecting) what events hold given a past path.

Finally, rules have the form $\varphi \leftarrow \psi$ and can be transaction or (complex) event rules. In a transaction rule $\varphi$ is a transaction atom and $\psi$ a transaction formula;  in an event rule  $\varphi$ is an event occurrence and $\psi$ is a event formula. A \emph{program} is a set of transaction and event rules.

Importantly, besides the data and transition oracles, \trev{} is also parametric on a $\fu{}$ function defining what event should be selected at a given time in case of conflict. Since defining what event should be picked from the set of occurring events depends on the application in mind, \trev{} does not commit to any particular definition, encapsulating it in function $\fu{}$.

As a reactive system, \trev{} receives a series of external events which may cause the execution of transactions in response. 
This is defined as $P,D_0 \seqDash \models e_1 \otimes \ldots \otimes e_k$, where $D_0$ is the initial KB state and $e_1 \otimes \ldots \otimes e_k$ is the sequence of external events that  arrive to the system.
Here, we want to find the path $D_0 \trans{O_1} \ldots \trans{O_n} D_n$ encoding a KB evolution that responds to $e_1 \otimes \ldots \otimes e_k$.
%

As mentioned, triggering explicit events is a transaction formula encoding the \emph{action} of making an occurrence explicitly true.
This is handled by the definition of interpretation, in a similar way to how atomic events defined by oracles primitives are made true:
\vspace{-.1cm}
\begin{definition}[\trev{} interpretations]
\label{def:interpretations-trev}
A \trev{} interpretation is a \tr{} interpretation that additionally satisfies the restriction: $3)\  \occ{e} \in M(\langle D\trans{\occ{e}}D \rangle)$ if $e \in \mathcal{P}_{e}$
\end{definition}
\vspace{-.1cm}

We can now define the satisfaction of complex formulas, and then models of a program.
Event formulas are evaluated w.r.t. the relation $\models_{\mathcal{TR}}$ specified in Def.~\ref{def:satev}.
Transaction formulas are evaluated w.r.t. the relation $\models$ which requires formulas to be true in \emph{expanded paths}, in which every occurring event is responded (something dealt by $\mathtt{exp}_{M}(\pi)$, defined  below).

\vspace{-.1cm}
\begin{definition}[Satisfaction of Transaction Formulas and Models]
\label{def:sat}
Let $M$ be an interpretation, $\pi$ a path, $\phi$ transaction formula. If $M(\pi) = \top$ then $M,\pi \models \phi$; else:
\vspace{-.2cm}
\begin{enumerate}[(iii)]
\item \textbf{Base Case:} $M,\pi \models p$ iff $\exists \pi'$ prefix of $\pi$ s.t.
$p \in M(\pi') $
and $\pi = \ex{\pi'}$, for every transaction atom $p$ where $p \not\in \mathcal{P}_{e}$.
\item \textbf{Event Case:} $M, \pi \models e$ iff $e \in \mathcal{P}_{e}$, $\exists \pi'$ prefix of $\pi$ s.t. $M,\pi'\models_{\mathcal{TR}} \occ{e}$ and $\pi = \ex{\pi'}$.
\item \textbf{Negation:} $M,\pi\models \neg \phi$ iff it is not the case that $M, \pi \models \phi$
\item \textbf{Disjunction:} $M,\pi \models \phi \lor \psi$ iff $M,\pi \models \phi$ or $M,\pi\models \psi$.
\item \textbf{Serial Conjunction:} $M,\pi \models \phi \otimes \psi$ iff $\exists \pi'$ prefix of $\pi$ and some split $\pi_{1}\circ\pi_{2}$ of $\pi'$ such that $M,\pi_1 \models \phi$ and $M,\pi_2\models \psi$ and $\pi = \ex{\pi'}$. 
\item \textbf{Executional Possibility:} $M,\pi \models \hyp \phi$ iff $\pi$ is a 1-path of the form $\langle D \rangle$ for some state $D$ and $M,\pi' \models \phi$ for some path $\pi'$ that begins at $D$.
\label{point-event}
\end{enumerate}

An interpretation $M$ is a \emph{model} of a transaction formula (resp. event formula) $\phi$ iff for every path $\pi$, $M,\pi \models \phi$ (resp. $M,\pi \models_{\mathcal{TR}} \phi$).
\noindent $M$ is a model of a program $P$ (denoted $M \models P$) iff it is a model of every (transaction and complex event) rule in $P$.
\end{definition}

 $\mathtt{exp}_{M}(\pi)$ is a function that, given a path with possibly unanswered events, expands it with the result of responding to those events. Its definition must perforce have some procedural nature: it must start by detecting which are the unanswered events; pick one of them, according to a given $\fu$ function; then expand the path with the response of the chosen event. The response to this event, computed by operator $\mathcal{R}_{M}$ defined below, may, in turn, generate the occurrence of further events. So, $\mathcal{R}_{M}$ must be iterated until no more unanswered events exist.
 
\begin{definition}[Expansion of a Path]
\label{def:expop}
\label{def:opresp}
For a path $\pi_{1}$ and an interpretation $M$, the response operator $\mathcal{R}_{M}(\pi_{1})$
is defined as follows:
$$
\mathcal{R}_{M}(\pi_{1}) = \left\{ 
\begin{array}{ll}
\pi_{1}\circ\pi_{2} &\text{ \textbf{if}   } \fu(M,\pi_{1}) = e \text{ and } M,\pi_{2} \models \resp{e} \\
\pi_{1} &\text{ \textbf{if}   } \fu(M,\pi_{1}) = \paempty %
\end{array} \right.
$$
The expansion of a path $\pi$ is $\ex{\pi} = \uparrow \mathcal{R}_{M}(\pi)$.
\end{definition}

In general it may not be possible to address all events in a finite path, and thus, $\mathcal{R}_{M}$ may not have a fixed-point.
In fact, non-termination is a known problem of reactive systems, and is often undecidable for the general case \cite{BaileyDR04}. 
However, if termination is possible, then a fixed-point exists and each iteration of $\mathcal{R}_{M}$ is an approximation of the expansion operator $\mathtt{exp}_{M}$.

%

This definition leaves open the $choice$ function, that is taken as a further parameter of \trev{}, and specifies how to choose the next unanswered event to respond to.
For its instantiation one needs to decide: 1) in which order should events be responded 
and 2) how should an event be responded.
The former  defines the handling order of events in case of conflict, e.g. based on when events have occurred (temporal order), on a priority list, or any other criteria.
The latter defines the response policy of an ECA-language, i.e. when is an event considered to be responded. E.g., if an event occurs more than once before the system can respond to it, this specifies if such response should be issued only once or equally to the amount of occurrences.
Choosing the appropriate operational semantics depends on the application in mind. 
In the following definition we exemplify how this $\fu{}$ function can be instantiated, for a case when events are responded in the (temporal) order in which they occurred, and events for which there was already a response are not responded again. 

%


\begin{definition}[Temporal $\fu$ function] 
Let $M$ be an interpretation and $\pi$ be a path. The temporal choice function is $\fu(M,\pi) =  \mathit{firstUnans}(M,\pi,order(M,\pi) )$ where:
\begin{itemize}
\item $order(M,\pi) = \langle e_1,\ldots,e_n\rangle$ iff $\forall e_i$ $1 \leq i \leq n$, $\exists \pi_i$ subpath of $\pi$ where $M,\pi \models_{\mathcal{TR}} \occ{e_i}$ and $\forall e_j$ s.t. $i< j$ then $e_j$ occurs after $e_i$
\item $e_{2}$ occurs after $e_{1}$ w.r.t. $\pi$ and $M$ iff 
there exists $\pi_{1},\pi_{2}$ subpaths of $\pi$ such that $\pi_{1} = \langle D_{i}\trans{O_{i}}\ldots\trans{O_{j-1}} D_{j} \rangle$, $\pi_{2} = \langle D_{n}\trans{O_{n}}\ldots\trans{O_{m-1}}D_{m} \rangle$,
$M,\pi_{1} \models \occ{e_{1}}$, $M,\pi_{2} \models \occ{e_{2}}$ 
and
$D_{j}\leq D_{m}$ w.r.t. the ordering in $\pi$.
\item $\mathit{firstUnans}(M,\pi,\langle e_1,\ldots, e_n \rangle) = e_i$
iff $e_i$ is the first event in  $\langle e_1,\ldots,e_n \rangle$ where
given $\pi'$ subpath of $\pi$ and $M,\pi' \models_{\mathcal{TR}} \occ{e}$ then $\neg \exists \pi''$ s.t. $\pi''$ is also a subpath of $\pi$, $\pi''$ is after $\pi'$ and $M,\pi'' \models \resp{e}$.
\end{itemize}
\end{definition}

We  continue by exemplifying the semantics in examples.

\begin{example}
\vspace{-0.5cm}
\begin{minipage}[r]{0.5\textwidth}
\small
\begin{equation}
\label{eq:1}
\tag{$P_3$}
\begin{array}{l}
\qquad \qquad p \leftarrow a.ins \\ 
\qquad \qquad \resp{e_{1}} \leftarrow c.ins
\end{array}
\end{equation}
\end{minipage}
\begin{minipage}{0.3\textwidth}
\small
\begin{equation}
\label{eq:2}
\tag{$P_4$}
\begin{array}{l}
p \leftarrow a.ins \\ 
\resp{e_{1}} \leftarrow c.ins\\
\occ{e_{1}} \leftarrow \occ{a.ins}
\end{array}
\end{equation}
\end{minipage}
\label{ex:1modtheory}

\noindent
Consider the programs\footnote{For brevity, in this and the following examples we assume the rule $\resp{p} \leftarrow \mathtt{true}$ to appear in every program for every primitive action $p$ defined in the signature of the oracles, unless when stated otherwise. I.e., we assume the responses of events inferred from primitive actions to hold trivially whenever their rules do not appear explicitly in the program.}  \ref{eq:1} and \ref{eq:2}.
In \ref{eq:1}, $p$ holds in the path $\langle \{\}\trans{\occ{a.ins}}\{a\} \rangle$. 
This is true since all interpretations must comply with the oracles and thus $\forall M$:  $a.ins \in M(\langle \{\}\trans{\occ{a.ins}}\{a\}\rangle)$ implying $M,\langle \{\}\trans{\occ{a.ins}}\{a\}\rangle \models a.ins$. Assuming that $M$ is a model of  \ref{eq:1}, then it satisfies the rule $p \leftarrow a.ins$, which means that $p \in M(\langle \{\}\trans{\occ{a.ins}}\{a\}\rangle)$ and $M,\langle \{\}\trans{\occ{a.ins}}\{a\} \rangle \models p$.

However, since $\occ{e_{1}} \leftarrow \occ{a.ins} \in$ \ref{eq:2} and $\forall M.\occ{a.ins} \in M(\langle \{\}\trans{\occ{a.ins}}\{a\} \rangle)$, for $M$ to be a model of \ref{eq:2},
then $\occ{e_{1}} \in M(\langle \{\}\trans{\occ{a.ins}}\{a\} \rangle)$.  Since $e_{1}$ has a response defined, then in path $\langle \{\}\trans{\occ{a.ins}}\{a\} \rangle$ the occurrence $e_{1}$ is unanswered and both the transactions $p$ and $a.ins$ cannot succeed in that path.
Namely, $\occ{e_{1}}$ constrains the execution of \emph{every} transaction in the path $\langle \{\}\trans{\occ{a.ins}}\{a\}\rangle$ and, for transaction formulas to succeed, such path needs to be \emph{expanded} with $e_{1}$'s response. 
Since, $\ex{\langle \{\}\trans{a.ins}\{a\} \rangle} = \langle \{\}\trans{\occ{a.ins}}\{a\}\trans{\occ{c.ins}}\{a,c\} \rangle$ then, both transactions $p$ and $a.ins$ succeed in the \emph{longer} path $\langle \{\}\trans{\occ{a.ins}}\{a\}\trans{\occ{c.ins}}\{a,c\} \rangle$, i.e. for an $M$ model of \ref{eq:2}:  $M,\langle\{\}\trans{\occ{a.ins}}\{a\}\trans{\occ{c.ins}}\{a,c\} \rangle \models p$ and $M,\langle\{\}\trans{\occ{a.ins}}\{a\}\trans{\occ{c.ins}}\{a,c\} \rangle \models a.ins$. Notice the non-monotonicity of \name{}, viz. that adding a new event rule to \ref{eq:1} falsifies the transaction formulas $p$ and $a.ins$ in paths where they were previously true.
\end{example}

As in \tr{}, in \trev{} every formula that is meant to be executed, is meant to be executed as a transaction.
As such, the primitive $a.ins$ in example~\ref{eq:2}  cannot succeed in the path $\langle \{\}\trans{\occ{a.ins}}\{a\} \rangle$ since there are unanswered events in that path.
However, note that $a.ins$  belongs to every interpretation $M$ of that path (due to the restrictions in Def.~\ref{def:interpretations}). Thus the primitive $a.ins$ is true in $\langle \{\}\trans{\occ{a.ins}}\{a\} \rangle$ although the transaction $a.ins$ is not.

\begin{example} 
\label{ex:1model} 
\vspace{-.7cm}
\mbox{}\\
\begin{minipage}{0.4\textwidth}
\small
$$
\begin{array}{l}
p \leftarrow a.ins \\
q \leftarrow b.ins \\
\resp{e_x} \leftarrow p \otimes q \\
\resp{e_{1}} \leftarrow d.ins \\
\resp{a.ins} \leftarrow c.ins \\
\occ{e_{1}}  \leftarrow \occ{a.ins} \otimes \occ{b.ins}  
\end{array}
$$
\end{minipage}
\begin{minipage}{0.4\textwidth}
\includegraphics[scale=0.75]{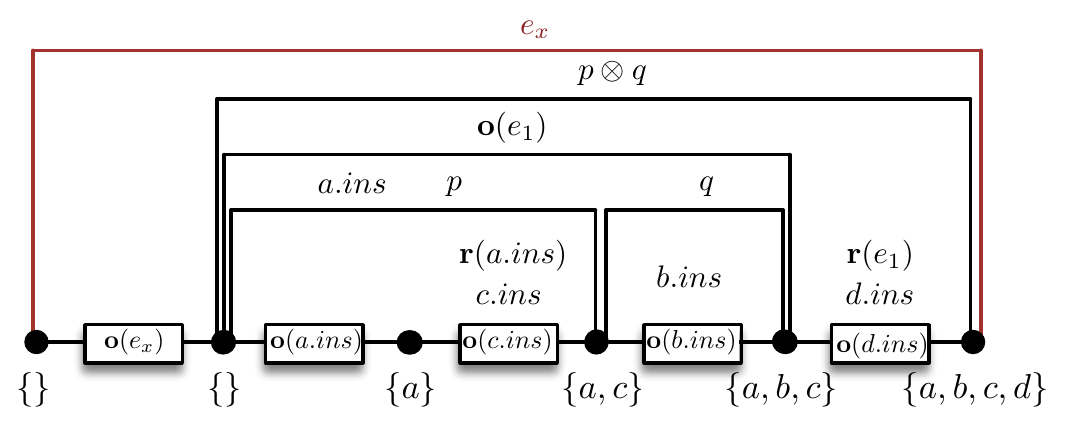}
\end{minipage}\\ 
The right-hand side figure illustrates a satisfaction of the external event $e_x$.
The occurrence of $e_x$ forces the satisfaction of the transaction $p \otimes q$, which is true if both its ``subformulas'' ($p$ and $q$) are satisfied over smaller paths. 
Note that, by definition of the relation $\models$, all occurrences 
detected over the independent paths that satisfy $p$ and $q$
are already responded in those paths. Thus, we need only to cater for the events triggered due to the serial conjunction.
Here, for a model $M$ of the program, $M,\langle \{\}\trans{\occ{a.ins}}\{a\}\trans{\occ{c.ins}}\{a,c\} \rangle \models p$ and $M,\langle \{a,c\}\trans{\occ{b.ins}}\{a,b,c\} \rangle \models q$. Further, the rule $ \occ{e_{1}} \leftarrow \occ{a.ins} \otimes \occ{b.ins} $ defines one pattern for the occurrence of $e_{1}$ which constrains the execution of transaction $p \otimes q$ and forces the expansion of the path to satisfy $\resp{e_1}$. Consequently,  $M,\langle\{\}\trans{\occ{a.ins}}\{a\}\trans{\occ{c.ins}}\{a,c\}\trans{\occ{b.ins}}\{a,b,c\}\trans{\occ{d.ins}}\{a,b,c,d\} \rangle  \models p \otimes q$, and  $M,\langle \{\} \trans{\occ{e_x}} \{\}\trans{\occ{a.ins}}\{a\}\trans{\occ{c.ins}}\{a,c\}\trans{\occ{b.ins}}\{a,b,c\}\trans{\occ{d.ins}}\{a,b,c,d\} \rangle \models e_x$
\end{example}

\vspace{-.3cm}
\subsection{Entailment and Properties}
\label{sec:entailment}

The logical entailment  defined  in Def.~\ref{model-tr}  can be used to reason about properties of transaction and event formulas that hold for \emph{every} possible path of execution. 
%
In \trev{}, similarly to \tr{}, we further define executional entailment, to talk about properties of a \emph{particular} execution path.
But, to reason about the execution of transactions over a specific path, care must be taken since, as described above, the satisfaction of a new occurrence in a path may invalidate transaction formulas that were previously true. 

To deal with a similar behavior, non-monotonic logics rely on the concept of minimal or preferred models:
instead of considering all possible models, non-monotonic theories restrict to the most skeptical ones. 
Likewise, \trev{} uses the minimal models of a program to define entailment, whenever talking about a particular execution of a formula.
As usual, minimality is defined by set inclusion on the amount of predicates that an interpretation satisfies, and a minimal model is a model that minimizes the set of formulas that an interpretation satisfies in a path.  

 
\begin{definition}[Minimal Model]
\label{def:minimalModels}
Let $M_{1}$ and $M_{2}$ be interpretations. Then $M_{1} \leq M_{2}$ if  $\forall \pi$: $ M_{2}(\pi) = \top \lor M_{1}(\pi) \subseteq M_{2}(\pi) $ \\
\noindent Let $\phi$ be a \name{} formula, and $P$ a program. $M$ is a \emph{minimal model} of $\phi$ (resp. $P$) if $M$ is a model of $\phi$ (resp. $P$) and $M \leq M'$ for every model $M'$ of $\phi$ (resp. $P$).
\end{definition}

Thus, to know if a formula succeeds in a particular path, we need only to consider the event occurrences \emph{supported} by that path, either because they appear as occurrences in the transition of states, or because they are a necessary consequence of the program's rules given that path.
Because of this, executional entailment in \trev{} is defined w.r.t. minimal models (cf. Def.~\ref{def:entailment}).

\begin{definition}[\trev{} Executional Entailment] 
\label{def:entailment2}
Let $P$ be a program, $\phi$ a transaction formula and $D_1\trans{O_{0}}\ldots\trans{O_{n}} D_n$ a path. 
Then $P, (D_1\trans{O_{0}}\ldots\trans{O_{n}} D_n) \models \phi$ ($\star$)
iff for every minimal model $M$ of $P$, $M,\langle D_1\trans{O_{0}}\ldots\trans{O_{n}} D_n \rangle \models \phi$. 
\noindent $P,D_1\seqDash \models \phi$
is said to be true, if there is a path $D_1\trans{O_{0}}\ldots\trans{O_{n}} D_n$ that makes ($\star$) true.
\end{definition}

\comment{
\vspace{-.2cm}
\begin{example} 
Recall program $P$ of example~\ref{ex:1model}.
For every minimal model $M_{m}$ of $P$ we have $M_{m}, \langle \{ \} \trans{\occ{e_x}} \{\}\trans{\occ{a.ins}}\{a\}\trans{\occ{c.ins}}\{a,c\}\trans{\occ{b.ins}}\{a,b,c\}\trans{\occ{d.ins}}\{a,b,c,d\} \rangle \models e_x$ and thus we know $P,(\{ \} \trans{\occ{e_x}} \{\}\trans{\occ{a.ins}}\{a\}\trans{\occ{c.ins}}\{a,c\}\trans{\occ{b.ins}}\{a,b,c\}\trans{\occ{d.ins}}\{a,b,c,d\}) \models e_x$
\end{example}
}

Interestingly, as in logic programs, formulas satisfied by this entailment have some support.
\begin{lemma}[Support]
\label{lemma:supported}
Let $P$ be a program, $\pi$ a path, $\phi$ a transaction atom. Then, if $P,\pi \models \phi$ one of the following holds:
\vspace{-.6cm}
\begin{enumerate}[(iii)]
\item $\phi$ is an elementary action and either $\phi \in \mathcal{O}^{d}(\pi)$ or $\phi \in \mathcal{O}^{t}(\pi)$;
\item $\phi$ is the head of a transaction rule in $P$ ($\phi \leftarrow body$) and $P,\pi \models body$;
\end{enumerate}
\end{lemma}
\extended{
\begin{proof}
Immediate by Definitions \ref{def:interpretations} and \ref{def:minimalModels}.
\end{proof}
}


As expected, \name{} extends \tr{}. Precisely, if a program $P$ has no complex event rules, and for every elementary action $a$ defined by the oracles the only rule for $\resp{a}$ in $P$ is $\resp{a} \leftarrow \tt{true}$, then executional entailment in \name{} can be recast in \tr{} if, \tr{} executional entailment is also restricted to minimal models. It is worth noting that, for a large class of \tr{} theories, and namely for the so-called serial-Horn theories, executional entailment in general coincides with that only using minimal models (cf. \cite{BonnerK93}). As an immediate corollary, it follows that if $P$ is \emph{event-free} and serial-Horn, then executional entailment in \name{} and in \tr{} coincide.

\vspace{-.3cm}
\section{Discussion and Related Work}
\label{sec:discussion}
Several solutions exist to reason about complex events.
Complex event processing (CEP) systems as \cite{snoopWindow,WuDR06} can reason efficiently with large streams of data and detect (complex) events.  These support a rich specification of events based on event pattern rules combining atomic events with some temporal constructs. 
As shown in Theorem~\ref{theorem:snoop}, \tr{} and \trev{} can express most event patterns of SNOOP and, ETALIS \cite{etalis} CEP system even uses \tr{}'s syntax and connectives, although abandoning \tr{}'s model theory and providing a different satisfaction definition. 
However, in contrast to \trev{}, CEP systems do not deal with the execution of actions in reaction to the events detected.

Extensions of Situation Calculus, Event Calculus, Action Languages, etc. exist with the ability to react to events, and have some transactional properties \cite{BaralLT97,BertossiPV98}. However, as in database triggers, these events are restricted to detect simple actions like ``on insert/delete" and thus have a very limited expressivity that fails to encode complex events, as defined in CEP systems and in \trev{}. 
%
%
To simultaneously reason about actions and complex events, ECA (following the syntax ``on \emph{event} if \emph{condition} do \emph{action}") languages \cite{AlferesBB11,BryEP06,ChomickiLN03} and logic programming based languages \cite{KowalskiS12,CostantiniG12} exist. 
%
These languages normally do not allow the action component of the language to be defined as a transaction, and when they do, they lack from a declarative semantics as \cite{PapamarkosPW06}; or they are based on active databases and can only detect atomic events defined as insertions/deletes \cite{Zaniolo95,LausenLM98a}.

In contrast, \name{} can deal with arbitrary atomic and complex events, and make these events trigger transactions. This is done by a logic-programming like declarative language. We have also defined a procedure to execute these reactive transactions, which is built upon the complex event detection algorithm of ETALIS and the execution algorithm of \tr{}, but is omitted for lack of space. 

\bibliographystyle{acmtrans}
\bibliography{bib}

\begin{thebibliography}{}

\bibitem[\protect\citeauthoryear{Adaikkalavan and Chakravarthy}{Adaikkalavan
  and Chakravarthy}{2004}]{snoopWindow}
{\sc Adaikkalavan, R.} {\sc and} {\sc Chakravarthy, S.} 2004.
\newblock Formalization and detection of events over a sliding window in active
  databases using interval-based semantics.
\newblock In {\em ADBIS}. 241--256.

\bibitem[\protect\citeauthoryear{Adaikkalavan and Chakravarthy}{Adaikkalavan
  and Chakravarthy}{2006}]{snoopIB}
{\sc Adaikkalavan, R.} {\sc and} {\sc Chakravarthy, S.} 2006.
\newblock Snoopib: Interval-based event specification and detection for active
  databases.
\newblock {\em Data Knowl. Eng.\/}~{\em 59,\/}~1, 139--165.

\bibitem[\protect\citeauthoryear{Alferes, Banti, and Brogi}{Alferes
  et~al\mbox{.}}{2011}]{AlferesBB11}
{\sc Alferes, J.~J.}, {\sc Banti, F.}, {\sc and} {\sc Brogi, A.} 2011.
\newblock Evolving reactive logic programs.
\newblock {\em Intelligenza Artificiale\/}~{\em 5,\/}~1, 77--81.

\bibitem[\protect\citeauthoryear{Anicic, Rudolph, Fodor, and Stojanovic}{Anicic
  et~al\mbox{.}}{2012}]{etalis}
{\sc Anicic, D.}, {\sc Rudolph, S.}, {\sc Fodor, P.}, {\sc and} {\sc
  Stojanovic, N.} 2012.
\newblock Stream reasoning and complex event processing in etalis.
\newblock {\em Semantic Web\/}~{\em 3,\/}~4, 397--407.

\bibitem[\protect\citeauthoryear{Bailey, Dong, and Ramamohanarao}{Bailey
  et~al\mbox{.}}{2004}]{BaileyDR04}
{\sc Bailey, J.}, {\sc Dong, G.}, {\sc and} {\sc Ramamohanarao, K.} 2004.
\newblock On the decidability of the termination problem of active database
  systems.
\newblock {\em Theor. Comput. Sci.\/}~{\em 311,\/}~1-3, 389--437.

\bibitem[\protect\citeauthoryear{Baral, Lobo, and Trajcevski}{Baral
  et~al\mbox{.}}{1997}]{BaralLT97}
{\sc Baral, C.}, {\sc Lobo, J.}, {\sc and} {\sc Trajcevski, G.} 1997.
\newblock Formal characterizations of active databases: Part ii.
\newblock In {\em DOOD}. LNCS, vol. 1341. Springer, 247--264.

\bibitem[\protect\citeauthoryear{Bertossi, Pinto, and Valdivia}{Bertossi
  et~al\mbox{.}}{1998}]{BertossiPV98}
{\sc Bertossi, L.~E.}, {\sc Pinto, J.}, {\sc and} {\sc Valdivia, R.} 1998.
\newblock Specifying active databases in the situation calculus.
\newblock In {\em SCCC}. IEEE Computer Society, 32--39.

\bibitem[\protect\citeauthoryear{Bonner and Kifer}{Bonner and
  Kifer}{1993}]{BonnerK93}
{\sc Bonner, A.~J.} {\sc and} {\sc Kifer, M.} 1993.
\newblock Transaction logic programming.
\newblock In {\em ICLP}. 257--279.

\bibitem[\protect\citeauthoryear{Bonner and Kifer}{Bonner and
  Kifer}{1998}]{BonnerK98b}
{\sc Bonner, A.~J.} {\sc and} {\sc Kifer, M.} 1998.
\newblock Results on reasoning about updates in transaction logic.
\newblock In {\em Transactions and Change in Logic Databases}. 166--196.

\bibitem[\protect\citeauthoryear{Bonner, Kifer, and Consens}{Bonner
  et~al\mbox{.}}{1993}]{BonnerKC93}
{\sc Bonner, A.~J.}, {\sc Kifer, M.}, {\sc and} {\sc Consens, M.~P.} 1993.
\newblock Database programming in transaction logic.
\newblock In {\em DBPL}. 309--337.

\bibitem[\protect\citeauthoryear{Bry, Eckert, and Patranjan}{Bry
  et~al\mbox{.}}{2006}]{BryEP06}
{\sc Bry, F.}, {\sc Eckert, M.}, {\sc and} {\sc Patranjan, P.-L.} 2006.
\newblock Reactivity on the web: Paradigms and applications of the language
  xchange.
\newblock {\em J. Web Eng.\/}~{\em 5,\/}~1, 3--24.

\bibitem[\protect\citeauthoryear{Chomicki, Lobo, and Naqvi}{Chomicki
  et~al\mbox{.}}{2003}]{ChomickiLN03}
{\sc Chomicki, J.}, {\sc Lobo, J.}, {\sc and} {\sc Naqvi, S.~A.} 2003.
\newblock Conflict resolution using logic programming.
\newblock {\em IEEE Trans. Knowl. Data Eng.\/}~{\em 15,\/}~1, 244--249.

\bibitem[\protect\citeauthoryear{Costantini and Gasperis}{Costantini and
  Gasperis}{2012}]{CostantiniG12}
{\sc Costantini, S.} {\sc and} {\sc Gasperis, G.~D.} 2012.
\newblock Complex reactivity with preferences in rule-based agents.
\newblock In {\em RuleML}. 167--181.

\bibitem[\protect\citeauthoryear{Kowalski and Sadri}{Kowalski and
  Sadri}{2012}]{KowalskiS12}
{\sc Kowalski, R.~A.} {\sc and} {\sc Sadri, F.} 2012.
\newblock A logic-based framework for reactive systems.
\newblock In {\em RuleML}. 1--15.

\bibitem[\protect\citeauthoryear{Lausen, Lud{\"a}scher, and May}{Lausen
  et~al\mbox{.}}{1998}]{LausenLM98a}
{\sc Lausen, G.}, {\sc Lud{\"a}scher, B.}, {\sc and} {\sc May, W.} 1998.
\newblock On active deductive databases: The statelog approach.
\newblock In {\em Transactions and Change in Logic Databases}. 69--106.

\bibitem[\protect\citeauthoryear{Papamarkos, Poulovassilis, and
  Wood}{Papamarkos et~al\mbox{.}}{2006}]{PapamarkosPW06}
{\sc Papamarkos, G.}, {\sc Poulovassilis, A.}, {\sc and} {\sc Wood, P.~T.}
  2006.
\newblock Event-condition-action rules on rdf metadata in p2p environments.
\newblock {\em Comp. Networks\/}~{\em 50,\/}~10, 1513--1532.

\bibitem[\protect\citeauthoryear{Wu, Diao, and Rizvi}{Wu
  et~al\mbox{.}}{2006}]{WuDR06}
{\sc Wu, E.}, {\sc Diao, Y.}, {\sc and} {\sc Rizvi, S.} 2006.
\newblock High-performance complex event processing over streams.
\newblock In {\em SIGMOD Conference}. ACM, 407--418.

\bibitem[\protect\citeauthoryear{Zaniolo}{Zaniolo}{1995}]{Zaniolo95}
{\sc Zaniolo, C.} 1995.
\newblock Active database rules with transaction-conscious stable-model
  semantics.
\newblock In {\em DOOD}. 55--72.

\end{thebibliography}

\end{document}